\documentclass[letterpaper]{article} % DO NOT CHANGE THIS
\usepackage{aaai24}  % DO NOT CHANGE THIS
\usepackage{times}  % DO NOT CHANGE THIS
\usepackage{helvet}  % DO NOT CHANGE THIS
\usepackage{courier}  % DO NOT CHANGE THIS
\usepackage[hyphens]{url}  % DO NOT CHANGE THIS
\usepackage{graphicx} % DO NOT CHANGE THIS
\usepackage{natbib}  % DO NOT CHANGE THIS AND DO NOT ADD ANY OPTIONS TO IT
\usepackage{caption} % DO NOT CHANGE THIS AND DO NOT ADD ANY OPTIONS TO IT
\usepackage{algorithm}
\usepackage{algorithmic}

\usepackage{newfloat}
\usepackage{listings}

\usepackage{subcaption}
\usepackage{color, colortbl}
\usepackage{multirow}
\usepackage{graphicx}
\usepackage{amsmath}
\usepackage{colortbl}
\usepackage{bm}
\usepackage{amssymb}
\usepackage{dsfont}

\usepackage{subcaption}
\usepackage{multirow}
\usepackage{graphicx}
\usepackage{amsmath}
\usepackage{colortbl}
\usepackage{bm}
\usepackage{booktabs}
\usepackage{array}
\usepackage{amssymb}

\DeclareMathOperator*{\argmin}{arg\,min}
\usepackage{theorem}
\newtheorem{definition}{Definition}
\newtheorem{mytheorem}{Theorem}
\usepackage{multicol}
\usepackage{pifont}

\DeclareCaptionStyle{ruled}{labelfont=normalfont,labelsep=colon,strut=off} % DO NOT CHANGE THIS
\floatstyle{ruled}
\newfloat{listing}{tb}{lst}{}
\floatname{listing}{Listing}
\newcommand{\myPara}[1]{\vspace{.05in}\noindent\textbf{#1}}

\title{M3D: Dataset Condensation by Minimizing Maximum Mean Discrepancy}
\author{
Hansong Zhang\textsuperscript{\rm 1, 2}\equalcontrib, Shikun Li\textsuperscript{\rm 1, 2}\equalcontrib, Pengju Wang\textsuperscript{\rm 1, 2}, Dan Zeng\textsuperscript{\rm 3}, Shiming Ge\textsuperscript{\rm 1, 2}\thanks{Corresponding Author}
}

\affiliations{
    \textsuperscript{\rm 1}Institute of Information Engineering, Chinese Academy of Sciences, Beijing 100092, China\\
    \textsuperscript{\rm 2}School of Cyber Security, University of Chinese Academy of Sciences, Beijing 100049, China\\
    \textsuperscript{\rm 3}Department of Communication Engineering, Shanghai University, Shanghai 200040, China\\
    \{zhanghansong,lishikun,wangpengju,geshiming\}@iie.ac.cn, dzeng@shu.edu.cn
}

\usepackage{bibentry}
\begin{document}

\maketitle

\begin{abstract}
Training state-of-the-art (SOTA) deep models often requires extensive data, resulting in substantial training and storage costs. To address these challenges, dataset condensation has been developed to learn a small synthetic set that preserves essential information from the original large-scale dataset. Nowadays, optimization-oriented methods have been the primary method in the field of dataset condensation for achieving SOTA results. However, the bi-level optimization process hinders the practical application of such methods to realistic and larger datasets. To enhance condensation efficiency, previous works proposed Distribution-Matching (DM) as an alternative, which significantly reduces the condensation cost. Nonetheless, current DM-based methods still yield less comparable results to SOTA optimization-oriented methods. In this paper, we argue that existing DM-based methods overlook the higher-order alignment of the distributions, which may lead to sub-optimal matching results. Inspired by this, we present a novel DM-based method named M3D for dataset condensation by \underline{\textbf{M}}inimizing the \underline{\textbf{M}}aximum \underline{\textbf{M}}ean \underline{\textbf{D}}iscrepancy  between feature representations of the synthetic and real images. By embedding their distributions in a reproducing kernel Hilbert space, we align all orders of moments of the distributions of real and synthetic images, resulting in a more generalized condensed set. Notably, our method even surpasses the SOTA optimization-oriented method IDC on the high-resolution ImageNet dataset. Extensive analysis is conducted to verify the effectiveness of the proposed method. Source codes are available at https://github.com/Hansong-Zhang/M3D.
\end{abstract}
\section{Introduction}
In the era of deep learning, the utilization of large-scale datasets comprising millions of samples has become an indispensable prerequisite for achieving state-of-the-art (SOTA) models~\cite{dsa, moderate_core}. However, the associated storage expenses and computational costs involved in training these models present formidable challenges, often rendering them beyond the reach of startups and non-profit organizations~\cite{dd_wang,highcost2,highcost3}.

% \begin{figure}[tb]
%     \centering
%     \includegraphics[width=0.45\textwidth]{figs/training_time.pdf}
%     \caption{Test accuracy (\%) trained on condensed ImageNet-10 images using ResNetAP-10 v.s. GPU hours to synthesize 10 images per class on a single RTX A6000 GPU.}
%     \label{train_time}
% \end{figure}

\begin{figure}[t]
  \centering
  \begin{subfigure}{0.32\linewidth}
    \includegraphics[width=\textwidth]{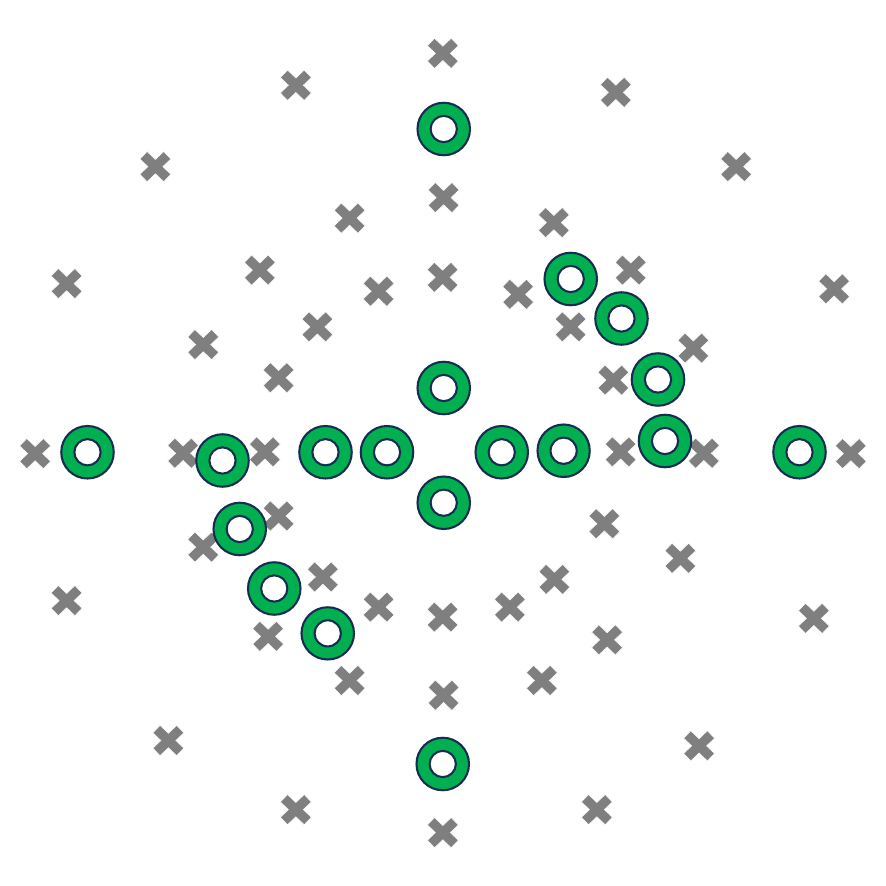}
    \caption{}
    \label{motivation1}
  \end{subfigure}
  \hfill
    \begin{subfigure}{0.32\linewidth}
    \includegraphics[width=\textwidth]{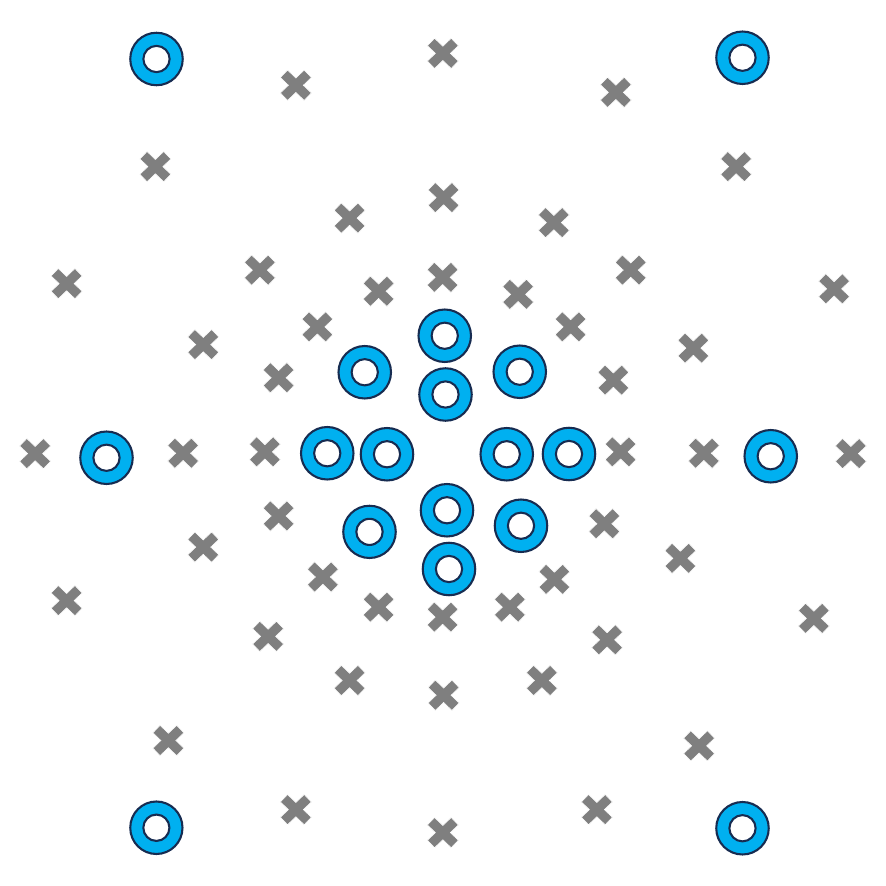}
    \caption{}
    \label{motivation2}
  \end{subfigure}
  \hfill
    \begin{subfigure}{0.32\linewidth}
    \includegraphics[width=\textwidth]{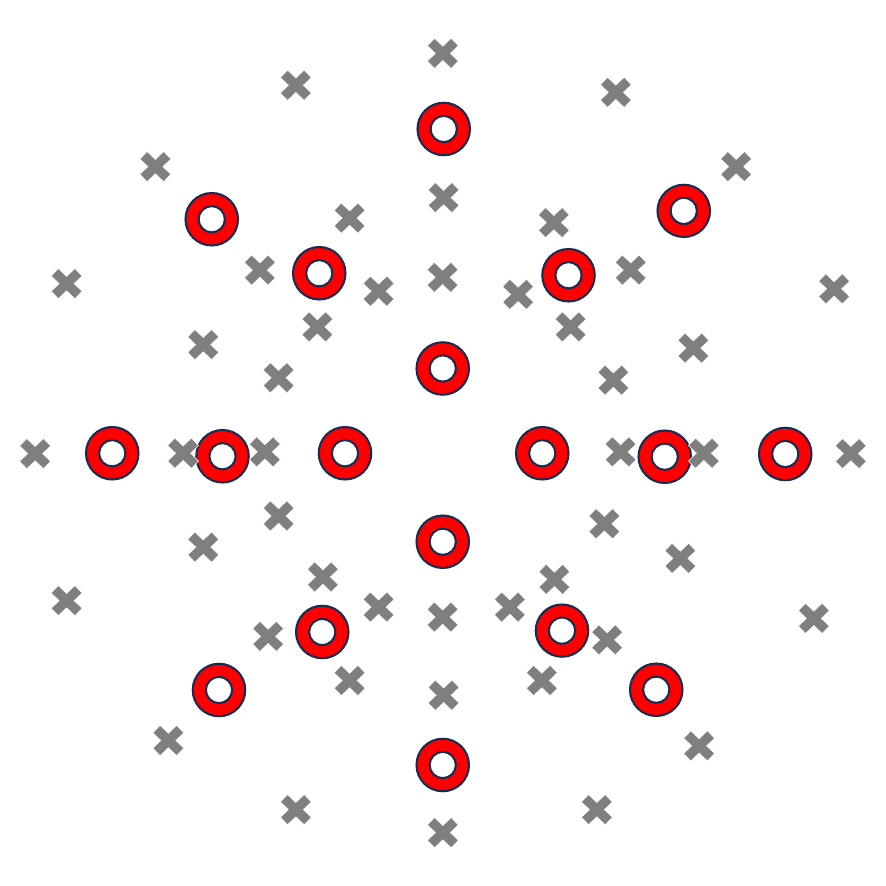}
    \caption{}
    \label{motivation3}
  \end{subfigure}
    \caption{Illustration of the importance of the higher-order alignment of distributions, where {circles} represent the representations of {synthesized} examples while {crosses} represent the representations of {original} examples. (a) The misaligned distributions with different second-order moments; (b) the misaligned distributions with different third-order moments; (c) the aligned distributions.}
  \label{motivation}
\end{figure}

To alleviate the challenges associated with larger datasets, Dataset Condensation (DC)~\cite{dd_wang} has emerged to reduce the training cost by synthesizing a compact set of informative images. Since its proposal, DC has attracted significant attention for addressing the challenges posed by the data burden~\cite{mtt,dc,idc,cafe,dm}. 
Typically, DC condenses the dataset by minimizing the distance between real and synthetic images via a pre-defined metric. Based on whether to perform a costly bi-level optimization~\cite{liu2021investigating}, these methods can be generally categorized into two groups: (1) \textit{Optimization-Oriented} methods~\cite{dc,idc,dsa,mtt}, which usually generate condensed examples by conducting performance matching or parameter matching via a bi-level optimization~\cite{yu2023dataset}; (2) \textit{Distribution-Matching(DM)-based} methods~\cite{cafe,dm}, which focus on aligning the feature distributions between real and synthetic data. Optimization-oriented methods have faced criticism for their inefficiency, primarily due to the involvement of bi-level optimization modules and time-consuming network updating processes~\cite{modelaugmentation,cafe,dm}. In contrast, DM-based methods do not involve such nested optimization of models, which significantly reduces the computational cost associated with dataset condensation. Nevertheless, the informativeness of the condensed examples generated by current DM-based methods may not be as comparable to those produced by optimization-oriented methods.

In this paper, we address a crucial oversight in existing DM-based methods, which is their neglect of higher-order moments of the distribution. As illustrated in Fig.~\ref{motivation}, despite sharing the same first moment, the representation distributions of original and synthetic examples with misaligned second-order moments (Fig.~\ref{motivation1}) or third-order moments (Fig.~\ref{motivation2}) can exhibit much distinct characteristics. Motivated by this issue, we propose a novel DM-based method involving \underline{\textbf{M}}inimizing the \underline{\textbf{M}}aximum \underline{\textbf{M}}ean \underline{\textbf{D}}iscrepancy (\textbf{M3D}) between the representation distributions of the real and synthetic images. Unlike previous DM-based methods that solely embed images in a feature representation space and align the first moment, our method further embeds the distribution of feature representations into a reproducing kernel Hilbert space. This transformation allows us to represent the infinite order of moments in a kernel-function form. By leveraging empirical estimation, we can readily align both first- and higher-order moments of the real and synthetic data with theoretical guarantees. 
Our method not only maintains the efficiency of the DM-based method but also exhibits significant improvements. Remarkably, the efficiency of our method makes it easily applicable to realistic and larger datasets like ImageNet~\cite{imagenet}.

Before delving into technical details, we clearly emphasize our contribution as:

\begin{itemize}
    \item We reveal the importance of the alignment of higher-order moments for distribution matching, which is overlooked by previous DM-based methods.
    \item We propose a theoretical-guaranteed method for dataset condensation named M3D, which applies the classical kernel method to represent an infinite number of moments in a kernel-function form, enabling the improved alignment of the higher-order moments of the representation distributions.
    \item We conduct extensive experiments to demonstrate the effectiveness and efficiency of our proposed method, where M3D yields SOTA performance with strong generalization across various scenarios.
\end{itemize}

\section{Background}
\myPara{Problem Fromulation.}
Dataset Condensation (DC)~\cite{dd_wang}, also called dataset distillation, targets to condense a large-scale dataset $\mathcal{T}=\{(\bm{x}_i, y_i)\}_{i=1}^{|\mathcal{T}|}$ into a tiny dataset $\mathcal{S}=\{(\bm{s}_j, y_j)\}_{j=1}^{|\mathcal{S}|}$, so that an arbitrary model trained on $\mathcal{S}$ achieves comparable performance to the one trained on $\mathcal{T}$. 
Typically, the condensed $\mathcal{S}$ is obtained by minimizing the information loss between the synthesized and the original examples, which can be formulated as:
\begin{equation}
    \mathcal{S}^\star = \argmin_\mathcal{S}\bm{D}(\phi(\mathcal{T}),\phi(\mathcal{S})),
\end{equation}
where $\bm{D}$ represents a distance metric such as {M}ean {S}quare {E}rror (MSE), and $\phi$ denotes the matching objective. 
As mentioned before, various objectives can lead to different optimization processes~\cite{yu2023dataset}, and based on whether to perform a costly bi-level optimization, existing methods can be mainly divided into optimization-oriented methods and Distribution-Matching (DM)-based methods. \footnote{Note that the introduction about more dataset condensation works can be found in the Appendix.}

\myPara{Distribution Matching.} Although optimization-oriented methods can achieve the SOTA performance, the inefficiency of them poses a significant obstacle to their application in realistic and larger datasets~\cite{modelaugmentation}. In response, DM-based methods have been developed as an alternative. In their pioneering work, DM~\cite{dm} introduces a surrogate matching objective that focuses on aligning the representation distributions of $\mathcal{S}$ and $\mathcal{T}$. This objective can be formulated as:
\begin{equation}
    \mathcal{S}^\star = \argmin_\mathcal{S}E_{\theta\sim P_{\theta}}\left[\bm{D}(g_{\theta}(\mathcal{S}),g_{\theta}(\mathcal{T}))\right],
\end{equation}
where $g_\theta$ is the deep encoder network parameterized as $\theta$, which is instanced by the model $f_\theta$ without the output layer. With MSE as the distance metric, the training objective of DM can be reformulated as:
\begin{equation}
    \mathcal{S}^\star = \argmin_\mathcal{S}E_{\theta\sim P_{\theta}}\lVert\frac{1}{|\mathcal{T}|}\sum_{i=1}^{|\mathcal{T}|}g_{\theta}(\bm{x}_i)-\frac{1}{|\mathcal{S}|}\sum_{j=1}^{|\mathcal{S}|}g_{\theta}(\bm{s}_j)\rVert^2,
    \label{dm}
\end{equation}
which works as minimizing the gap between empirical first moment of the representation distributions between $\mathcal{S}$ and $\mathcal{T}$. 
Compared to previous optimization-oriented methods, DM~\cite{dm} eliminates the need for network updating, relying instead on randomly initialized encoders. Furthermore, the costly bi-level optimization is avoided in DM, leading to significantly improved training efficiency.

\myPara{\textit{Remark.}} Given the lower effectiveness of DM compared to optimization-oriented SOTA methods, efforts have been made to enhance DM and generate more informative examples in previous works~\cite{idm, sajedi2023datadam}. For instance, IDM~\cite{idm} enhances DM through techniques such as partitioning, enriched model sampling, and class-aware regularization. Similarly, DataDAM~\cite{sajedi2023datadam} improves DM by incorporating attention matching. In contrast to these methods where only the first-order moment is matched, our focus is on enhancing DM through distribution embedding and higher-order moments, which are also noticed but not addressed explicitly by IDM~\cite{idm}.

\myPara{Reproducing Kernel Hilbert Space.} We provide a brief recap of the Reproducing Kernel Hilbert Space (RKHS)~\cite{muandet2017kernel,smola2007hilbert,borgwardt2006integrating} here, which serves as the foundation of our method.
\begin{definition}
  Given a kernel $\mathcal{K}$, $\mathcal{H}$ is a Hilbert space of functions $\mathcal{X}\rightarrow \mathbb{R}$ with dot product $\langle\cdot,\cdot\rangle$, if $\forall \phi$, satisfying the reproducing property:
  \begin{equation}
      \langle \phi(\cdot),\mathcal{K}(x,\cdot)\rangle=\phi(x).
       \label{hilbert}
  \end{equation}
\end{definition}
That is to say, with the RKHS, we can map a function $f$ on $\mathcal{X}$ to its value at $x$ as an inner product. In addition to the reproducing property mentioned above, the kernel function $\mathcal{K}$ must also satisfy the following two properties:
\begin{equation*}
\begin{aligned}
     \textbf{Symmetry}:& \quad\mathcal{K}(x,x')=\mathcal{K}(x',x)\\
    \textbf{Positive}:& \quad\mathcal{K}(\cdot,\cdot) \geq 0 
\end{aligned}
\end{equation*}
Commonly used kernel function include the polynomial kernel $\mathcal{K}(x,x')= (x^\intercal x'+c)^d$, the Gaussian RBF kernel $\mathcal{K}(x,x')=\exp{(-\lambda\lVert x-x'\rVert^2)}$, and the Linear kernel $\mathcal{K}(x,x')=x^\intercal x'$. 

\section{Methodology}
In this section, we begin by analyzing the importance of the alignment of higher-order moments for distribution matching. Subsequently, we propose our method M3D by exploiting the classical kernel method~\cite{empirical_MMD,borgwardt2006integrating} to align the higher-order moments of the representation distributions between real and synthesized data with theoretical guarantees.
\subsection{Importance of the Higher-Order Alignment}
As shown in Eq.~(\ref{dm}), it is evident that DM~\cite{dm} only considers aligning the first moment
(mean) of the representation distributions, while neglecting higher-order moments. At a high level, it may lead to the higher-order misalignment of the representation distributions of its condensed data and original data.

To investigate this misalignment issue and highlight the importance of the higher-order alignment, we assessed the moment distances between the condensed set and the original training set on CIFAR-10 with 10 images per class. This was done by incorporating higher-order moment regularization terms into the original loss of DM~\cite{dm}.
The results, presented in Table~\ref{explore}, reveal that adding second-order regularization notably decreases the distance between higher-order moments of the condensed and original data, underscoring the inadequacy of aligning only the first moment. Furthermore, performing more regularization enhances the condensed dataset's performance through improved higher-order alignment. These results underscore the critical role of higher-order moment alignment in distribution matching, which is neglected in previous works.

\begin{table}[tb]
    \centering
    \small
    \setlength\tabcolsep{1.6pt}
     % \resizebox{0.46\textwidth}{!}{
    \begin{tabular}{lccc|c}
    \toprule
         & 1st-order & 2nd-order & 3rd-order  & \multirow{2}{*}{Test Acc.}\\
         & (mean) & (variance) & (skewness)  & \\
         \midrule
        DM & 4.91 & 7.37 & 6.76 &   48.9\\
        +2nd {Reg.} & 4.51 & 6.96 & 6.35 & 52.1 ({$\uparrow$ 3.2})\\
        +2nd $\&$ 3rd {Reg.} & 3.69 & 6.14 & 5.94 & 53.9 ({$\uparrow$ 5.0})\\
        \midrule
        \textbf{M3D} & \textbf{0.82} & \textbf{1.23} &  \textbf{1.64}  & \textbf{63.5} ({$\uparrow$ 14.6})\\
        \bottomrule
    \end{tabular}
    % }
    \caption{{The distance between the moments of the condensed set and the original training set.} ``+2nd(3rd) {Reg.}'' denotes adding the regularization of aligning the 2nd(3rd)-order moment to the original loss of DM.}
    \label{explore}
\end{table}

\subsection{Minimizing Maximum Mean Discrepancy}
\begin{figure*}
    \centering
\includegraphics[width=\textwidth]{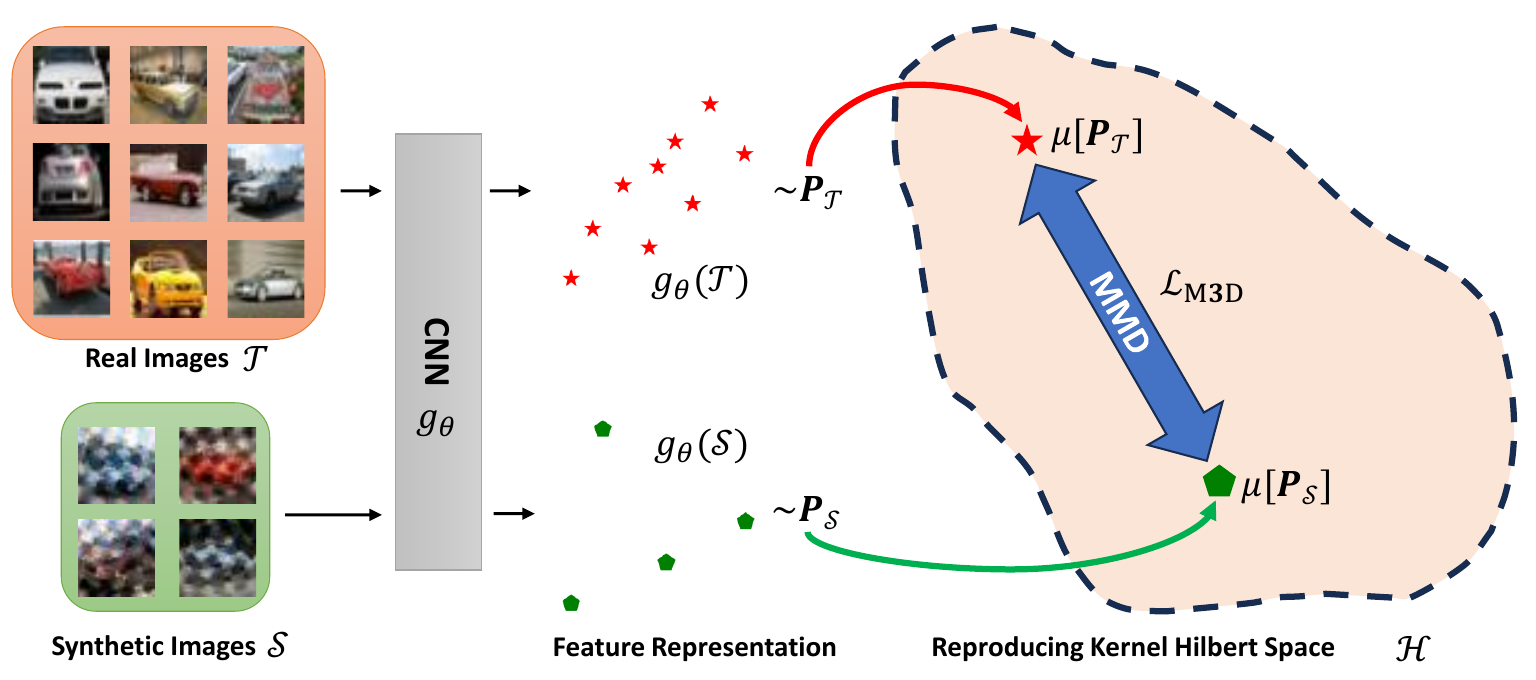}
    \caption{The framework of M3D. After extracting the representations via a encoder network, the distributions of real and synthetic representations are further embedded in the Reproducing Kernel Hilbert Space (RKHS), where the M3D loss $\mathcal{L}_{\text{M3D}}$ is calculated to guild the update of synthetic examples for higher-order distribution alignment.}
     \label{framework}
\end{figure*}

From the preceding analysis, it becomes evident that perfecting distribution matching necessitates the consideration of higher-order moments. While incorporating higher-order regularizations directly aids in aligning these moments, it is limited to finite moments. Moreover, tuning the regularization coefficient becomes increasingly challenging with a growing number of regularization terms. In this subsection, we represent a new DM-based method that aligns the infinite order of moments in a kernel-function form. We depict the framework of the proposed M3D in Fig.~\ref{framework}.

\myPara{Embedding Distribution in RKHS.}~
Denoting the distribution of representations for real and synthetic examples as $g_\theta(\mathcal{T})\sim \bm{P}_\mathcal{T}$ and $g_\theta(\mathcal{S})\sim \bm{P}_\mathcal{S}$ respectively, where $g_\theta$ denotes the representation extractor parameterized by $\theta$.  As the order of moments extends infinitely, it is impractical to explicitly align an infinite number of moments. To address this, we need to first embed the distribution in an RKHS $\mathcal{H}$:
\begin{equation}
\mu[\bm{P}_{\mathcal{T}/\mathcal{S}}]:=E_{\mathcal{T}/\mathcal{S}}[\mathcal{K}(g_\theta(\bm{x}/\bm{s}),\cdot)],
\label{app_deriv}
\end{equation}
which has been proven to be a valid embedding for distance based on the following theorem:
\begin{mytheorem}~\cite{injective}
  If the kernel function $\mathcal{K}$ is universal, then the mean map $\mu:=\bm{P}\rightarrow \mu[\bm{P}]$ is injective.
\end{mytheorem}

\myPara{Maximum Mean Discrepancy.} Via the reproducing property of $\mathcal{H}$, $\forall \phi$, we have
\begin{equation}
\langle \phi,\mu[\bm{P}_{\mathcal{T}/\mathcal{S}}]\rangle = E_{\mathcal{T}/\mathcal{S}}[\phi(g_\theta(\bm{x}/\bm{s}))],
\end{equation}
which indicate that we can compute expectations w.r.t. $\bm{P}_{\mathcal{T}/\mathcal{S}}$ by taking the inner product with the distribution kernel embedding $\mu[\bm{P}_{\mathcal{T}/\mathcal{S}}]$. This property is favorable because it helps us to calculate the {M}aximum {M}ean {D}iscrepancy (MMD) between $\bm{P}_\mathcal{T}$ and $\bm{P}_\mathcal{S}$:
\begin{equation*}
\begin{aligned}
        \text{MMD}(\bm{P}_\mathcal{T}, \bm{P}_\mathcal{S}):&=\sup(E_\mathcal{T}[\phi(g_\theta(\bm{x}))] - E_\mathcal{S}[\phi(g_\theta(\bm{s}))]) \\ 
  & = \sup\langle\phi,\mu[\bm{P}_\mathcal{T}]-\mu[\bm{P}_\mathcal{S}]\rangle,
\end{aligned}
\end{equation*}
where $\phi\in\mathcal{H}$ and $\lVert\phi\rVert_\mathcal{H}\leq 1$. In addition, based on the Cauchy-Schwarz inequality, we have $\langle\phi,\mu[\bm{P}_\mathcal{T}]-\mu[\bm{P}_\mathcal{S}]\rangle\leq \lVert\phi\rVert_\mathcal{H} \lVert\mu[\bm{P}_\mathcal{T}]-\mu[\bm{P}_\mathcal{S}]\rVert_\mathcal{H}\leq\lVert\mu[\bm{P}_\mathcal{T}]-\mu[\bm{P}_\mathcal{S}]\rVert_\mathcal{H}$, hence the MMD can be further simplified as:
\begin{equation}
        \text{MMD}(\bm{P}_\mathcal{T}, \bm{P}_\mathcal{S})=\lVert\mu[\bm{P}_\mathcal{T}]-\mu[\bm{P}_\mathcal{S}]\rVert.
        \label{mmd1}
\end{equation}
It should be noted that $\mu[\bm{P}_\mathcal{T}]$ and $\mu[\bm{P}_\mathcal{S}]$ are characterized by infinite-dimensional spaces, which renders direct computation unattainable. However, we can leverage the reproducing property of the RKHS to transform them into a more tractable form using the kernel function $\mathcal{K}$. This transformation can be formally expressed as:
\begin{equation}
    \text{MMD}^2(\bm{P}_\mathcal{T}, \bm{P}_\mathcal{S})=\mathcal{K}_{\mathcal{T,T}}+\mathcal{K}_{\mathcal{S,S}}-2\mathcal{K}_{\mathcal{T,S}},
    \label{mmd2}
\end{equation}
where $\mathcal{K}_{X,Y}=E_{X,Y}[\mathcal{K}(g_\theta(x), g_\theta(y))]$ with $x\sim X, y\sim Y$. Due to limited page, we provide the derivation of Eq.~(\ref{mmd2}) in the Appendix. Last, note that we only have access to the datasets $\mathcal{T}$ and $\mathcal{S}$ rather than their underlying distributions. In order to tackle this issue, denoting the empirical approximation of $\mu[\bm{P}_{\mathcal{T}}]$ and $\mu[\bm{P}_{\mathcal{S}}]$ as $\mu[\mathcal{T}]=\frac{1}{|\mathcal{T}|}\sum_{i=1}^{|\mathcal{T}|}\mathcal{K}(g_\theta(\bm{x}_i),\cdot)$, $\mu[\mathcal{S}]=\frac{1}{|\mathcal{S}|}\sum_{j=1}^{|\mathcal{S}|}\mathcal{K}(g_\theta(\bm{s}_j),\cdot)$ respectively, we introduce the following theorem:
\begin{mytheorem}~\cite{empirical_MMD}
\label{theo2}
    Assume that $\lVert\phi\rVert_\infty\leq R$ for all $\phi\in\mathcal{H}$ with $\lVert\phi\rVert_\mathcal{H}\leq 1$. Then with probability at least $1-\delta$, $\lVert\mu[\bm{P}_{\mathcal{T/S}}]-\mu[\mathcal{T/S}]\rVert\leq 2\bar{R}(\mathcal{H}, \bm{P}_{\mathcal{T}/\mathcal{S}})+R\sqrt{-|\mathcal{T}/\mathcal{S}|^{-1}\log(\delta)}$, where $\bar{R}(\mathcal{H}, \bm{P}_{\mathcal{T}/\mathcal{S}})$ is the Rademacher average which is ensured to yield error of $\mathcal{O}(\sqrt{|\mathcal{T}/\mathcal{S}|^{-1}})$.
\end{mytheorem}
Theorem~\ref{theo2} guarantees that the empirical approximations $\mu[\mathcal{T/S}]$ are good proxies for $\mu[\bm{P}_{\mathcal{T/S}}]$. Therefore, we can modify Eq.~(\ref{mmd2}) to the following empirical form as the M3D loss:
\begin{equation}
\label{emp_mmd}
    \mathcal{L}_{\text{M3D}}=\hat{\text{MMD}}^2(\bm{P}_\mathcal{T}, \bm{P}_\mathcal{S})=\hat{\mathcal{K}}_{\mathcal{T,T}}+\hat{\mathcal{K}}_{\mathcal{S,S}}-2\hat{\mathcal{K}}_{\mathcal{T,S}},
\end{equation}
where $\hat{\mathcal{K}}_{X,Y}=\frac{1}{|X|\cdot|Y|}\sum_{i=1}^{|X|}\sum_{j=1}^{|Y|}\mathcal{K}(g_\theta(x_i), g_\theta(y_j))$ with $\{x_i\}_{i=1}^{|X|}\sim X, \{y_j\}_{j=1}^{|Y|}\sim Y.$ 
Based on the analysis above, we have successfully achieved the transformation of an infinite number of moments into a finite form using RKHS. As shown in Table~\ref{explore}, this transformation allows us to effectively align the distributions between $\mathcal{T}$ and $\mathcal{S}$ during the condensing process. 

\subsection{Training Algorithm of M3D}
The pseudo-code of M3D is provided in the Appendix. In addition to the kernel method, we exploit the following two techniques to enhance the distribution matching.

\myPara{Factor $\&$ Up-sampling.}~The factor technique~\cite{idc}, also termed as partitioning and expansion augmentation in IDM~\cite{idm}, aims to increase the number of representations extracted from $\mathcal{S}$ without additional storage cost. Specifically, with the factor parameter being $l$, each image $\bm{s}_i \in \mathcal{S}$ is factorized into $l\times l$ mini-examples and then up-sampled to its original size in training:
\begin{equation}
    \bm{s}_i\xrightarrow{\text{Factor}}
    \left[
\begin{matrix}
  \bm{s}_i^{1,1}  & \dots & \bm{s}_i^{1,l}\\
  \vdots & \ddots & \vdots\\
  \bm{s}_i^{l,1} & \dots & \bm{s}_i^{l,l}
\end{matrix}
\right]
\xrightarrow{\text{Up-sample}}
\{\bm{s}_i^{'1}, \bm{s}_i^{'2},\dots,\bm{s}_i^{'l\times l}\}.
\end{equation}
In this way, the storage space of $\mathcal{S}$ can be further leveraged. Following previous works, the same factor technique is incorporated into our framework, where we further exploit its benefits in aligning distributions in higher-order moments.

\myPara{Iteration per Random Model.} Following DM~\cite{dm}, we employ multiple randomly initialized models to extract representation embeddings from both $\mathcal{T}$ and $\mathcal{S}$. In contrast to DM, where only a single-step iteration is performed for each model, we posit that relying solely on the representation distributions of one batch of real and synthetic examples may introduce matching biases. To address this, without incurring additional memory usage, we empirically observe that conducting multiple iterations per model (IPM) enhances the performance of the condensed set.

\section{Experiments}
\begin{figure*}[tb]
    \centering
    \includegraphics[width=\textwidth]{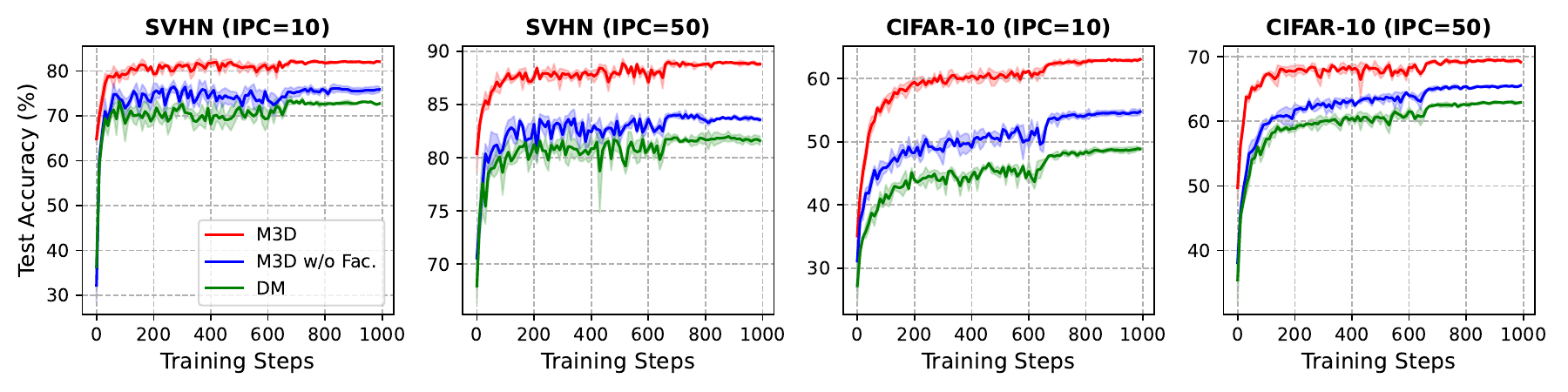}
    \caption{{Performance comparison between M3D and DM across varying training steps.} M3D w/o Fac denotes the M3D without using the factor technique.}
    \label{loss_curve}
\end{figure*}

\begin{table*}[tb]
    \centering
    \small
    \setlength\tabcolsep{1.5pt}
    % \resizebox{\textwidth}{!}{
    \begin{tabular}{ccc|ccc|ccccccc|c}
    \toprule
       \multirow{2}{*}{Dataset}  &  \multirow{2}{*}{IPC} & \multirow{2}{*}{Ratio $(\%)$} & \multicolumn{3}{c|}{Coreset Selection} & \multicolumn{7}{c|}{Dataset Condensation} & \multirow{2}{*}{Whole}\\
                  &  &  & Random & Herding & K-Center &  DC & DSA & CAFE & CAFE+DSA & DM & IDM & \textbf{M3D} & \\\midrule
    \multirow{3}{*}{MNIST} & 1 & 0.017 & 64.9$_{\pm3.5}$ & 89.2$_{\pm1.6}$ & 89.3$_{\pm1.5}$  & 91.7$_{\pm0.5}$ & 88.7$_{\pm0.6}$ & 93.1$_{\pm0.3}$ & 90.8$_{\pm0.5}$ & 89.7$_{\pm0.6}$ & - & \textbf{94.4$_{\pm0.2}$}  & \multirow{3}{*}{99.6$_{\pm0.0}$}  \\
                           & 10 & 0.17 & 95.1$_{\pm0.9}$ & 93.7$_{\pm0.3}$ & 84.4$_{\pm1.7}$ & 97.4$_{\pm0.2}$ & \textbf{97.8$_{\pm0.1}$} & 97.2$_{\pm0.2}$ & 97.5$_{\pm0.1}$ & 97.5$_{\pm0.1}$ & - &{97.6$_{\pm0.1}$} &    \\
                           & 50 & 0.83 & 97.9$_{\pm0.2}$ & 94.8$_{\pm0.2}$ & 97.4$_{\pm0.3}$ & 98.8$_{\pm0.2}$ & \textbf{99.2$_{\pm0.1}$} & 98.6$_{\pm0.2}$ & {98.9$_{\pm0.2}$} & 98.6$_{\pm0.1}$ & - & 98.2$_{\pm0.2}$ &   \\\midrule
    \multirow{3}{*}{F-MNIST} & 1 & 0.017 & 51.4$_{\pm3.8}$ & 67.0$_{\pm1.9}$ & 66.9$_{\pm1.8}$ & 70.5$_{\pm0.6}$ & 70.6$_{\pm0.6}$ & {77.1$_{\pm0.9}$} & 73.7$_{\pm0.7}$ & 70.7$_{\pm0.6}$$^\dagger$  & - & \textbf{80.7$_{\pm0.3}$} & \multirow{3}{*}{93.5$_{\pm0.1}$}  \\
                           & 10 & 0.17 & 73.8$_{\pm0.7}$ & 71.1$_{\pm0.7}$ & 54.7$_{\pm1.5}$ &  82.3$_{\pm0.4}$ & {84.6$_{\pm0.3}$} & 83.0$_{\pm0.4}$ & 83.0$_{\pm0.3}$ & 83.5$_{\pm0.3}$$^\dagger$  & - & \textbf{85.0$_{\pm0.1}$} &   \\
                           & 50 & 0.83 & 82.5$_{\pm0.7}$ & 71.9$_{\pm0.8}$ & 68.3$_{\pm0.8}$ & 83.6$_{\pm0.4}$ & \textbf{88.7$_{\pm0.2}$} & 84.8$_{\pm0.4}$ & {88.2$_{\pm0.3}$} & 88.1$_{\pm0.6}$$^\dagger$  & - & 86.2$_{\pm0.3}$ &   \\\midrule
    \multirow{3}{*}{SVHN} & 1 & 0.014 & 14.6$_{\pm1.6}$ & 20.9$_{\pm1.3}$ & 21.0$_{\pm1.5}$ & 31.2$_{\pm1.4}$ & 27.5$_{\pm1.4}$ & 42.6$_{\pm3.3}$ & {42.9$_{\pm3.0}$} & 30.3$_{\pm0.1}$$^\dagger$  & - & \textbf{62.8$_{\pm0.5}$} & \multirow{3}{*}{95.4$_{\pm0.1}$}  \\
                           & 10 & 0.14 & 35.1$_{\pm4.1}$ & 50.5$_{\pm3.3}$ & 14.0$_{\pm1.3}$ & 76.1$_{\pm0.6}$ & {79.2$_{\pm0.5}$} & 75.9$_{\pm0.6}$ & 77.9$_{\pm0.6}$ & 73.5$_{\pm0.5}$$^\dagger$  & - & \textbf{83.3$_{\pm0.7}$} &   \\
                           & 50 & 0.7 & 70.9$_{\pm0.9}$ & 72.6$_{\pm0.8}$ & 20.1$_{\pm1.4}$ & 82.3$_{\pm0.3}$ & {84.4$_{\pm0.4}$} & 81.3$_{\pm0.3}$ & 82.3$_{\pm0.4}$ & 82.0$_{\pm0.2}$$^\dagger$  & - & \textbf{89.0$_{\pm0.2}$} &   \\\midrule
    \multirow{3}{*}{CIFAR-10} & 1 & 0.02 & 14.4$_{\pm2.0}$ & 21.5$_{\pm1.2}$ & 21.5$_{\pm1.3}$ & 28.3$_{\pm0.5}$ & 28.8$_{\pm0.7}$ & 30.3$_{\pm1.1}$ & {31.6$_{\pm0.8}$} & 26.0$_{\pm0.8}$  & \textbf{45.6$_{\pm0.7}$} & 45.3$_{\pm0.3}$ & \multirow{3}{*}{84.8$_{\pm0.1}$}  \\
                           & 10 & 0.2 & 26.0$_{\pm1.2}$ & 31.6$_{\pm0.7}$ & 14.7$_{\pm0.9}$ & 44.9$_{\pm0.5}$ & {52.1$_{\pm0.5}$} & 46.3$_{\pm0.6}$ & 50.9$_{\pm0.5}$ & 48.9$_{\pm0.6}$  & 58.6$_{\pm0.1}$ & \textbf{63.5$_{\pm0.2}$} &   \\
                           & 50 & 1 & 43.4$_{\pm1.0}$ & 40.4$_{\pm0.6}$ & 27.0$_{\pm1.4}$ & 53.9$_{\pm0.5}$ & 60.6$_{\pm0.5}$ & 55.5$_{\pm0.6}$ & 62.3$_{\pm0.4}$ & {63.0$_{\pm0.4}$} & 67.5$_{\pm0.1}$  & \textbf{69.9$_{\pm0.5}$} &   \\\midrule
    \multirow{3}{*}{CIFAR-100} & 1 & 0.2 & 4.2$_{\pm0.3}$ & 8.4$_{\pm0.3}$ & 8.3$_{\pm0.3}$ &  12.8$_{\pm0.3}$ & 13.9$_{\pm0.3}$ & 12.9$_{\pm0.3}$ & {14.0$_{\pm0.3}$} & 11.4$_{\pm0.3}$  & 20.1$_{\pm0.3}$ & \textbf{26.2$_{\pm0.3}$} & \multirow{3}{*}{56.2$_{\pm0.3}$}  \\
                           & 10 & 2 & 14.6$_{\pm0.5}$ & 17.3$_{\pm0.3}$ & 7.1$_{\pm0.2}$ & 25.2$_{\pm0.3}$ & {32.3$_{\pm0.3}$} & 27.8$_{\pm0.3}$ & 31.5$_{\pm0.2}$ & 29.7$_{\pm0.3}$ & \textbf{45.1$_{\pm0.1}$} & {42.4$_{\pm0.2}$} &   \\
                           & 50 & 10 & 30.0$_{\pm0.4}$ & 33.7$_{\pm0.5}$ & 30.5$_{\pm0.3}$ & - & 42.8$_{\pm0.4}$ & 37.9$_{\pm0.3}$ & 42.9$_{\pm0.2}$ & {43.6$_{\pm0.4}$} & 50.0$_{\pm0.2}$ & \textbf{50.9$_{\pm0.7}$} &   \\
    \bottomrule
    \end{tabular}
    % }
    \caption{{Comparison with previous coreset selection and dataset condensation methods on low-resolution datasets.} All the datasets are condensed using a 3-layer ConvNet. IPC: image(s) per class. Ratio $(\%)$: the ratio of condensed examples to the whole training set. ``$\dagger$'' denotes the result is reproduced by us. Best results are in {bold}. Note that some entries are marked as ``-'' because of scalability issues or the results are not reported.}
    \label{results}
\end{table*}

\begin{table}[tb]
    \centering
    \small
    \setlength\tabcolsep{2.7pt}
    % \resizebox{0.45\textwidth}{!}{
    \begin{tabular}{l|cccc|cccc}
    \toprule
         & \multicolumn{4}{c|}{ImageNet-10} &  \multicolumn{4}{c}{ImageNet-100}\\
        IPC & \multicolumn{2}{c}{10} & \multicolumn{2}{c|}{20} & \multicolumn{2}{c}{10} & \multicolumn{2}{c}{20}\\
        \midrule
        &   Acc. & Time & Acc. & Time &   Acc. & Time & Acc. & Time \\ \midrule
      Random   & 46.9 & - & 51.8 & - & 20.7 & - & 29.7 & -\\
      Herding   & 50.4 & - & 57.5 & -   & 22.6 & -  &31.1 & - \\
      \midrule
    DSA   & 52.7  & 27.0h  & 57.4 & 51.4h  & 21.8 & 9.7h   &  30.7&23.9h  \\
      IDC   &  72.8 & 70.1h &  76.6 & 92.8h  &  46.7 & 141.0h  &  53.7 & 185.0h  \\
      DM   &  52.3 & {1.4h}  &  59.3 & 3.6h  &  22.3 & \textbf{2.8h}   &  30.4&\textbf{2.8h}   \\
      \textbf{M3D}   &  \textbf{73.4} & \textbf{1.1h}  &  \textbf{76.8} & \textbf{3.1h}  &  \textbf{46.9} &3.5h &  \textbf{55.5} & 4.2h\\
      \bottomrule
    \end{tabular}
    % }
    \caption{{The performance and efficiency comparison on high-resolusion ImageNet-subsets}.  The synthetic examples are condensed using ResNetAP-10. The minimal time required for obtaining the best performance is reported, which is measured on a single RTX-A6000 GPU with same batch size. For ImageNet-100, all methods are splitted into five sub-tasks with 20 classes each for faster optimization.}
    \label{imagenet_results}
\end{table}

\begin{figure*}[tb]
\centering
\begin{subfigure}[b]{0.33\textwidth}
    \includegraphics[width=\textwidth]{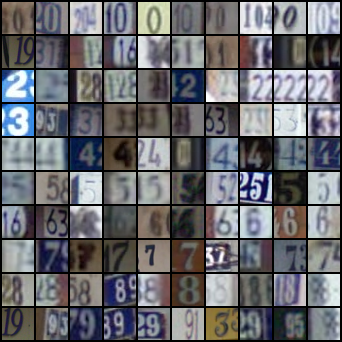}
    \caption{Initialized SVHN images.}
\end{subfigure}
\begin{subfigure}[b]{0.33\textwidth}
    \includegraphics[width=\textwidth]{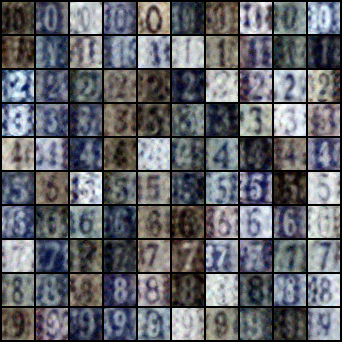}
    \caption{Condensed images by {DM}.}
\end{subfigure}
\begin{subfigure}[b]{0.33\textwidth}
    \includegraphics[width=\textwidth]{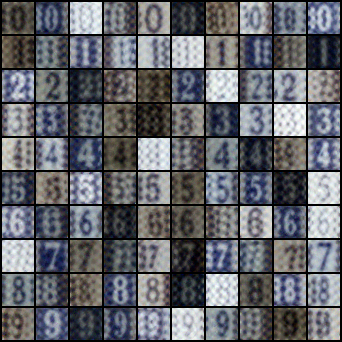}
    \caption{Condensed images by {M3D}.}
\end{subfigure}
\caption{{Visualization of the condensed set of SVHN dataset with 10 images per class.} The condensed set is generated by (b) DM and (c) M3D. Both DM and M3D use the same initialization as (a) shows.}
\label{svhn_vis}
\end{figure*}

In this section, we begin by comparing our proposed M3D with SOTA baselines on multiple benchmark datasets. Subsequently, we conduct an in-depth examination of M3D through ablation analysis.
\subsection{Experimental Setups}
\myPara{Datasets.}~We evaluate the classification performance of networks trained on synthetic images that have been condensed using various baselines as well as our proposed method M3D. Our evaluation encompasses five low-resolution datasets: MNIST~\cite{mnist}, Fashion-MNIST (F-MNIST)~\cite{fmnist}, SVHN~\cite{svhn}, CIFAR-10~\cite{cifar}, and CIFAR-100~\cite{cifar}. In addition, we also conduct experiments on the high-resolution dataset ImageNet subsets~\cite{imagenet}. Detailed descriptions of datasets can be found in the Appendix.

\myPara{Network Architectures.}~We use a depth-3 ConvNet~\cite{convnet} for the low-resolution datasets, and a ResNetAP-10~\cite{idc} (ResNet-10 with the strided convolution replaced by average pooling) for the high-resolution ImageNet subsets.

\myPara{Baselines.}~We employ an extensive range of methods as baselines for comparison. Regarding coreset selection methods, we consider the following: (1) Random, (2) Herding~\cite{core_herding}, and (3) K-Center~\cite{core_kcenter1,core_kcenter2}. For optimization-oriented DC methods, we evaluate (4) DC~\cite{dc}, (5) DSA~\cite{dsa}, (6) IDC~\cite{idc}. On the other hand, for DM-based DC methods, we include (7) CAFE~\cite{cafe}, (8) its variant CAFE+DSA~\cite{cafe}, (9) DM~\cite{dm} and (10) IDM~\cite{idm}.  We provide detailed descriptions of baselines in the Appendix.

\myPara{Metric.} Following previous works~\cite{dd_wang,dm,idc}, we employ the test accuracy of networks trained on condensed examples as the evaluation metric. All the networks are trained from scratch for multiple times — 10 times for low-resolution datasets and 3 times for ImageNet subsets. We report the average performance and the standard deviation.

\myPara{Implementation Details.}~We employ the Gaussian kernel for RKHS by default. The number of iterations is set to 10K for all low-resolution datasets. While for ImageNet subsets, we set 1K iterations. Additionally, the number of iterations per model is consistently set to 5 across all datasets. Regarding the learning rates for the condensed data, we assign a value of 1 for low-resolution datasets including F-MNIST, SVHN and CIFAR-10/100. For ImageNet subsets, we adopt a learning rate of 1e-1. Following IDC~\cite{idc}, the factor parameter $l$ is set to 2 for low-resolution datasets and 3 for ImageNet subsets.

\subsection{Comparison to the SOTA Methods}
% \myPara{Results on Low-resolution Datasets.}~
Table~\ref{results} and Table~\ref{imagenet_results} present the comparison of our method with coreset selection and dataset condensation methods. The results show that synthetic examples are more informative than the selected ones, especially when the number of image(s) per class is small. 
This is attributed to the fact that synthetic examples are not confined to the set of real examples. Furthermore, our method consistently outperforms other baselines across a diverse set of scenarios. Remarkably, M3D achieves over a 5$\%$ higher accuracy than the best baseline on SVHN, CIFAR-10 (IPC=10), and CIFAR-100 (IPC=1). Notably, for high-resolution ImageNet subsets~\cite{imagenet,idc,modelaugmentation}, our method surpasses all baselines in test accuracy, including the current SOTA optimization-oriented IDC~\cite{idc}. It is worth noting that IDC~\cite{idc} demands an exceptionally long time to condense ImageNet subsets, e.g., approximately 4 days on ImageNet-10 with IPC=20~\cite{modelaugmentation}. In contrast, M3D achieves superior performance in a matter of hours. Additionally, our method eliminates the need for network updates, thereby circumventing the tuning of various hyper-parameters. Consequently, our method can be readily applied to realistic and larger datasets, maintaining efficiency and effectiveness simultaneously.

To further demonstrate the advantages of our method, we provide the test accuracy across varying training steps in Fig.~\ref{loss_curve}. As observed, our method consistently outperforms DM at different training steps. Even without the factor technique, our method still achieves considerable improvement, highlighting the effectiveness of M3D in aligning distributions compared to previous DM-based methods.

\myPara{Cross-Architecture Evaluation.}~
We further assess the performance of our condensed examples on different architectures. In Table~\ref{cross_arch}, we present the performance of our condensed examples from  CIFAR-10 dataset on ConNet-3, ResNet-10~\cite{resnet10}, and DenseNet-121~\cite{densenet}. Combining the results from Table~\ref{results}, we can find that M3D outperforms the compared methods not only on the architecture used for condensation but on unseen ones.

\begin{figure}[tb]
\centering
    \includegraphics[width=0.45\textwidth]{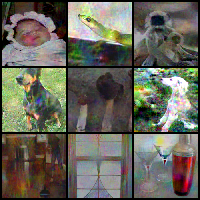}
\caption{{Representative samples condensed by M3D on ImageNet.} The corresponding labels, from left to right and top to bottom, are bonnet, green snake, langur, doberman, gyromitra, saluki, vacuum, window screen, and cockroach.}
\label{imagenet_vis}
\end{figure}

\myPara{Visualizations.}
We visualize the condensed images of SVHN and ImageNet in Fig.~\ref{svhn_vis} and Fig.~\ref{imagenet_vis}, respectively. For SVHN, we initialize the synthetic set $\mathcal{S}$ using random images from the training set $\mathcal{T}$ and then apply the condensation process using DM and M3D. As shown, the condensed images by DM and M3D appear as if the original images have been augmented with a distinct texture. Notably, the condensed images produced by our method exhibit a more pronounced and visually appealing texture compared to DM. While the overall appearance remains similar, our condensed images demonstrate better alignment with the higher-order moments of the original training set. In the case of ImageNet, the condensed images exhibit a texture reminiscent of a sunspot. In contrast to optimization-oriented methods, the images condensed by M3D retain more natural features and are more visually recognizable to humans. More visualization results are provided in the Appendix.

\begin{figure}[tb]
  \centering
    \begin{subfigure}{0.49\linewidth}
    \includegraphics[width=\textwidth]{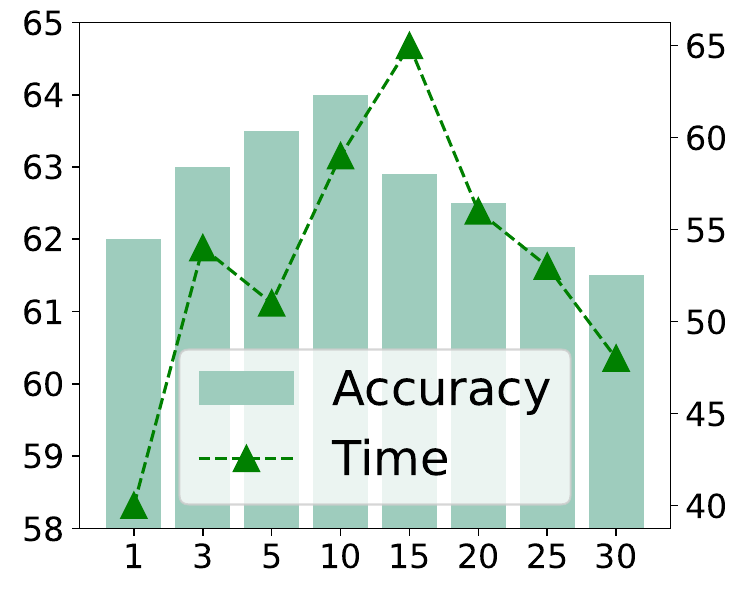}
    \caption{}
    \label{abinner}
  \end{subfigure}
  \hfill
      \begin{subfigure}{0.49\linewidth}
    \includegraphics[width=\textwidth]{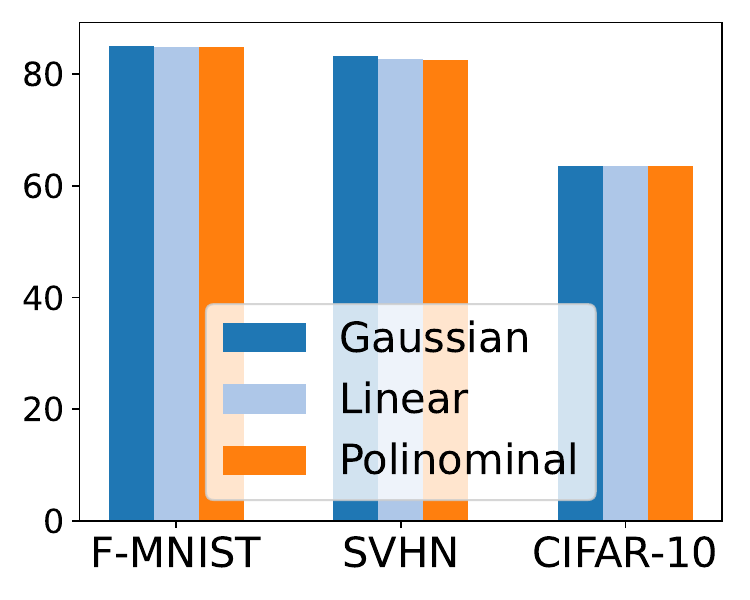}
    \caption{}
    \label{abkernel}
  \end{subfigure}
  \caption{(a) Ablation of IPM, where the horizontal axis represents the number of IPM, the left and right vertical axis denote test accuracy ($\%$) and the corresponding time cost (mins), respectively. (b) Ablation of the kernel function, where the vertical axis denotes test accuracy ($\%$).}
    % \caption{(a) Ablation of the number of the iteration per model (IPM), where the ``Time'' denotes the minimal time cost to get the corresponding accuracy and the experiments are conducted on CIFAR-10 with 10 images per class; (b) ablation of the kernel function.}
  \label{ablation_fig}
\end{figure}

\begin{table}
    \centering
    \small
    % \resizebox{0.35\textwidth}{!}{
    \begin{tabular}{cc|ccc}
    \toprule
     \multirow{2}{*}{IPC}    & Evaluation & \multicolumn{3}{c}{Method}\\
     \cline{3-5}
      & Model & DSA & DM & M3D \\
      \midrule
      \multirow{3}{*}{10}    & ConvNet-3 & \underline{52.1} & 48.9 & \textbf{63.5}\\
         & ResNet-10 & 32.9 & \underline{42.3} & \textbf{56.7}\\
         & DenseNet-121 & 34.5 & \underline{39.0} & \textbf{54.6} \\
              \midrule
      \multirow{3}{*}{50}    & ConvNet-3 & 60.6 &\underline{63.0} & \textbf{69.9}\\
         & ResNet-10 & 49.7  & \underline{58.6} & \textbf{66.6}\\
         & DenseNet-121 & 49.1 & \underline{57.4} & \textbf{66.1} \\
              \midrule
    \end{tabular}
    % }
    \caption{{Cross-architecture generalization performance ($\%$) on CIFAR-10.} The synthetic examples is condensed using ConvNet-3 and evaluated using other architectures.}
    \label{cross_arch}
\end{table}

\subsection{Ablation Study}

\myPara{Impact of the Iteration per Model (IPM).}~
We conduct experiments using various number of iterations per model, and the corresponding performance is depicted in Fig.~\ref{abinner}. We adopt CIFAR-10 with 10 images per class to showcase the impact of IPM. In addition to the test accuracy of condensed examples, we also provide the training time required to achieve the reported accuracy. As shown, increasing the number of IPM may lead to improved performance of the condensed data, but it also increases the training time. Conversely, an excessively large IPM can compromise the generalization ability of the condensed examples.

\myPara{Impact of the Kernel Function.}~
Different kernel functions construct distinct Reproducing Kernel Hilbert Spaces (RKHS). To investigate their influence, we adopt two additional kernel functions in addition to the Gaussian kernel: the linear kernel and the polynomial kernel. Fig.~\ref{abkernel} illustrates the test accuracy under different kernel functions with 10 images per class. As observed, the choice of $\mathcal{K}$ has minimal impact on the performance of the condensed dataset. This indicates that as long as the selected kernel function is valid, our M3D can effectively embed the distributions in the constructed RKHS, resulting in a robust method.

\section{Conclusion}
In conclusion, this paper introduces a novel Distribution-Matching (DM)-based method called M3D for dataset condensation. 
With a theoretical guarantee, our method embeds the representation distributions of real and synthetic examples in a reproducing kernel Hilbert space, minimizing the maximum mean discrepancy between them to align their distributions in both first- and higher-order moments.
Extensive experiments show the effectiveness and efficiency of our method. Notably, the efficiency of our method enables its application to more realistic and larger datasets. 
M3D first studies the alignment of higher-order moments of the representation distributions between real and synthetic examples, and establishes a strong baseline in DM-based methods for dataset condensation, which we believe will be valuable to the research community.

\section*{Acknowledgments}
This work was partially supported by grants from the National Key Research and Development Plan (2020AAA0140001), and the Beijing Natural Science Foundation (19L2040), and the Open Research Project of National  Key Laboratory of Science and Technology on Space-Born Intelligent Information Processing (TJ-02-22-01).

\nocite{*}
\bibliography{aaai24}

\clearpage
\appendix

\onecolumn
\section{Pseudo-Code of M3D}\label{sec:psudocode}
The pseudo-code of M3D is presented in Algorithm~\ref{pseudocode}.
\begin{algorithm}[h]
    \caption{M3D: Dataset Condensation by Minimizing Maximum Mean Discrepancy}
    \label{algori}
    \textbf{Input}: Original training dataset $\mathcal{T}$\\
    \textbf{Required}: Randomly initialized synthetic set $\mathcal{S}$, the learning rate of synthetic images $\eta$, the number of total iterations $I_{\text{total}}$, the number of iteration(s) per model $I_{\text{pm}}$. Denote the number of classes as $C$.\\
    \textbf{Output}: Condensed dataset $\mathcal{S^\star}$
    \begin{algorithmic}[1] %[1] enables line numbers
    \FOR{$i=0,...,I_{\text{total}}-1$}
    \IF{$i \% I_{\text{pm}}=0$}
    \STATE Randomly initialize a network $g_{\theta_i}$
    \ENDIF
    \FOR{$c=0,...,C-1$}
    \STATE Sample a mini-batch of both real and synthetic set in class $c$: $\mathcal{B}^\mathcal{T}_c\sim \mathcal{T}$ and $\mathcal{B}^\mathcal{S}_c\sim \mathcal{S}$
    \STATE Obtain the representations of the real and synthetic images $g_\theta(\mathcal{B}^\mathcal{T}_c)$ and $g_\theta(\mathcal{B}^\mathcal{S}_c)$
    \STATE Compute the M3D loss $\mathcal{L}_{\text{M3D}}$ between the sampled mini-batches according to Eq.~(\ref{emp_mmd})
    \STATE Update $\mathcal{S}$ as: $\mathcal{S} = \mathcal{S} - \eta\nabla_{\mathcal{S}}\mathcal{L}_{\text{M3D}}$
    \ENDFOR
    \ENDFOR
    
    \STATE \textbf{return} $\mathcal{S}$ as $\mathcal{S^\star}$
    \end{algorithmic}
    \label{pseudocode}
\end{algorithm}

\section{Related Works}
\label{related_work}
\textit{Dataset Condensation} aims to obtain a synthetic dataset that are valid to replace original one in downstream training. According to different optimization objectives, current works in \textit{Dataset Condensation} can be roughly divided into two categories: \textbf{Optimization-oriented} methods and \textbf{Distribution-mathcing-based} methods. Next, we will elaborate these two categories in detail:
\subsection{Optimization-Oriented Methods}~
Optimization-oriented methods adopt a bi-level optimization fashion to learn the synthetic dataset~\cite{mtt,dc,idc,dsa,modelaugmentation,du2023minimizing,dd_wang}.
The pioneering works by Wang \textit{et al.}~\cite{dd_wang} in \textit{Dataset Condensation} poses a strong assumption that a model trained on the condensed dataset should be identical to that trained on the original dataset. However, directly aligning the converged models proves impractical due to the vast parameter space and convergence challenges. Consequently, subsequent studies adopt a more stringent assumption that the two models should follow a similar optimization path. To realize this objective, they leverage techniques such as gradient matching or aligning training trajectories of synthetic and real images. 
However, though the models can be aligned along the optimization path, these methods still can not bypass the time-consuming bi-level optimization process, which hinders their application in practical scenarios.
Therefore, we refer to them as optimization-oriented methods, which can be further classified into two mainstream solutions: performance matching and parameter matching.

\myPara{Performance Matching.}~Performance matching is initially introduced by Wang \textit{et al.}~\cite{dd_wang}, wherein the synthetic dataset is trained in a manner that ensures the model trained on it achieves the lowest loss on the original dataset. In this way, the performance of models trained on synthetic and real dataset could be matched. Subsequent work by Dong \textit{et al.}~\cite{deng2022remember} proposes that substituting the model updating with the inner loop with momentum can improve the performance of synthetic images. Further, to mitigate the inefficiency of the meta-gradient backpropagation, Nguyen \textit{et al.}~\cite{nguyen2020dataset} replace the nerural network in the inner loop with a kernel model. With kernel ridge regression (KRR), the synthetic dataset can be updated by backpropagating meta-gradient through the kernel function~\cite{lei2023comprehensive,nguyen2020dataset,KIP,zhou2022dataset,loo2022efficient,loo2023dataset}, which greatly reduces the training cost. Following the KRR stream, Jacot \textit{et al.}~\cite{jacot2018neural} proposed a theory which proves the equivalance between KRR and training infinite-width neural networks. Based on this, Nguyen \textit{et al.}~\cite{KIP} adopts infinite-width neural networks as the kernel function to dataset condensation, which establishes a close relationship between KRR and deep learning. Further, a similar method was also proposed by Zhou \textit{et al.}~\cite{zhou2022dataset}, which also focuses on the last-layer in neural networks with KRR.

\myPara{Parameter Matching.}~
The methods of parameter matching in dataset condensation is first proposed by Zhao \textit{et al.}~\cite{dc}, which has been extended to various following works~\cite{dsa,mtt,idc}. The key idea of parameter matching is that: the parameter induced by real and synthetic datasets should be consistent with each other. In DC~\cite{dc}, the resultant gradients w.r.t. the model on the synthetic dataset are encouraged to be close to that on the real dataset. DSA~\cite{dsa} further proposes to add the differentiable siamese augmentation before the images are feed to the model, which improved the informativeness of the synthetic dataset. Recognizing that single-step gradients may accumulate errors, MTT~\cite{mtt} introduces a multi-step parameter matching method. This approach involves iteratively updating the synthetic images to ensure that the model trained on them follows similar trajectories towards convergence. 
On the other hand, there are some works focusing on effective model updating during the condensing process. The original work DC~\cite{dc} uses the synthetic dataset to update the network, which could cause the overfitting problem in the eraly stage of training. To tackle this, IDC~\cite{idc} proposes to update the model by real dataset, in which way the overfitting problem can be alleviated due to the large size of real dataset. Moreover, IDC~\cite{idc} also proposed the factor technique, which can enrich the information of synthetic dataset by factoring and up-sampling. Though effective, these parameter matching methods are computationally expensive, as thousands of differently initialized networks are required to update the synthetic dataset. To accelerate the condensing process, \cite{modelaugmentation} propose model augmentation, which adds a Gaussian perturbation on the early-stage models to save the time and storage cost to condense dataset.

\subsection{Distribution-Matching-based Methods}~
The goal of distribution-matching-based methods is to obtain synthetic dataset which shares similar feature distribution with real dataset~\cite{sajedi2023datadam,dm,idm,cafe}. Different from optimization-oriented methods, distribution-matching-based methods do not need the bi-level optimization or the meta-gradient to condense the dataset, which greatly reduces the time and memory cost of the condensing process. DM~\cite{dm} condenses the dataset by aligning the representation embedding of real and synthetic images. Besides, it discard the model updating process, as the trained models have little impact on the performance of the synthetic images. Similarly, CAFE~\cite{cafe} aligns the embedding of not only the last layer but also the frontier layers to learn the synthetic dataset.
Moreover, with the discriminant loss term, CAFE~\cite{cafe} can enhance the discriminative property of the synthetic dataset. To further improve distribution matching, IDM~\cite{idm} proposes the ``partitioning and expansion'' technique to increase the number of representations extracted from the synthetic dataset. Besides, IDM~\cite{idm} also uses the trained model to condense the dataset, with a cross-entropy regularization loss, IDM successfully alleviates DM~\cite{dm}'s class misalignment issue. In addition to aligning the representation distribution, DataDAM~\cite{sajedi2023datadam} adds the spatial attention matching to enhance the performance of the synthetic set.
Although previous distribution-matching-based methods have shown promising results in  efficiency, they all 
%adopt the empirical Maximum Mean Discrepancy (MMD) as the distance between the synthetic and real datasets, which can
only align the first-order moment of the feature distributions of synthetic and real dataset. As a result, the higher-order moment of the distributions may be misaligned by previous methods. To this end, in this paper we propose to embed the image representations to a reproducing kernel Hilbert space (RKHS), where we can simultaneously align each moment of the real and synthetic image representations, making the representation distribution of synthetic dataset more consistent with the real one.
\subsection{Coreset Selection}~
Instead of synthesizing data, coreset selection methods~\cite{core_herding,core_kcenter1,core_forgetting,core_kcenter2,moderate_core,yang2022dataset} select a subset of the whole training set based on a pre-defined criterion. For instance, Herding~\cite{core_herding} selects samples that are close to the class centers; K-center~\cite{core_kcenter1,core_kcenter2} selects multiple center points of a class to minimize the maximum distance between the selected samples and their nearest center. However, the performance of the coreset can not be guaranteed because the criterion is heuristic. Moreover, the coreset is restricted by the quality of the original images, which further hinders their application to reduce the data burden.

\section{Derivation of MMD}\label{derivation}
Recall Eq.~(\ref{mmd1}) in the main manuscript that: 
\begin{equation*}
        \text{MMD}(\bm{P}_\mathcal{T}, \bm{P}_\mathcal{S})=\lVert\mu[\bm{P}_\mathcal{T}]-\mu[\bm{P}_\mathcal{S}]\rVert.
\end{equation*}
Squaring both sides, we have
\begin{equation*}
\begin{aligned}
    &\text{MMD}^2(\bm{P}_\mathcal{T}, \bm{P}_\mathcal{S})=\lVert\mu[\bm{P}_\mathcal{T}]-\mu[\bm{P}_\mathcal{S}]\rVert^2=\lVert\mu[\bm{P}_\mathcal{T}]\rVert^2+\lVert\mu[\bm{P}_\mathcal{S}]\rVert^2-2\lVert\mu[\bm{P}_\mathcal{T}]\rVert\lVert\mu[\bm{P}_\mathcal{T}]\rVert \\
    &=E^2_{\mathcal{T}}[\mathcal{K}(g_\theta(\bm{x}),\cdot)]+E^2_{\mathcal{S}}[\mathcal{K}(g_\theta(\bm{s}),\cdot)]-E_{\mathcal{T}}[\mathcal{K}(g_\theta(\bm{x}),\cdot)]E_{\mathcal{S}}[\mathcal{K}(g_\theta(\bm{s}),\cdot)] \,\,//\textcolor{blue}{\text{Substituting Eq.~(\ref{app_deriv})}}\\
    &=E_{\mathcal{T},\mathcal{T}}[\langle\mathcal{K}(g_\theta(\bm{x}),\cdot),\mathcal{K}(g_\theta(\bm{x}),\cdot) \rangle]+E_{\mathcal{S},\mathcal{S}}[\langle\mathcal{K}(g_\theta(\bm{s}),\cdot),\mathcal{K}(g_\theta(\bm{s}),\cdot) \rangle]-2E_{\mathcal{T},\mathcal{S}}[\langle\mathcal{K}(g_\theta(\bm{x}),\cdot),\mathcal{K}(g_\theta(\bm{s}),\cdot) \rangle]\\
    &=E_{\mathcal{T},\mathcal{T}}[\mathcal{K}(g_\theta(\bm{x}),g_\theta(\bm{x}))]+E_{\mathcal{S},\mathcal{S}}[\mathcal{K}(g_\theta(\bm{s}),g_\theta(\bm{s}))]-2E_{\mathcal{T},\mathcal{S}}[\mathcal{K}(g_\theta(\bm{x}),g_\theta(\bm{s}))]
\,\,//\textcolor{blue}{\text{Reproducing property of $\mathcal{K}$}}
\end{aligned}
\end{equation*}

\section{Dataset Description}\label{detailed_data_description}
We evaluate our method on four low-resolution datasets and one high-resolution dataset.

\subsection{Low-Resolution Datasets}
\begin{itemize}
    \item \textbf{Fashion-MNIST}. Fashion-MNIST~\cite{fmnist} is a popular dataset commonly used for evaluating machine learning algorithms. It contains 60,000 training images and 10,000 testing images, all in grayscale with a size of 28x28 pixels. The dataset comprises 10 different fashion categories, including items like T-shirts, dresses, and shoes.
    \item \textbf{SVHN}. SVHN~\cite{svhn} is a large-scale dataset primarily designed for digit recognition in natural images. It consists color images containing house numbers captured from Google Street View. The dataset is divided into two subsets: a training set with 73,257 images, a test set with 26,032 images. SVHN is commonly used for tasks like digit localization and recognition in real-world scenarios.
    \item \textbf{CIFAR-10/100}. CIFAR-10 and CIFAR-100~\cite{cifar} are widely used benchmark datasets for object recognition and classification tasks. CIFAR-10 consists of 60,000 color images, with 50,000 images for training and 10,000 images for testing. The dataset covers 10 different object classes, including common objects like cars, birds, and cats. On the other hand, CIFAR-100 contains 100 object classes, with 600 images per class. Each image in both datasets has a size of 32x32 pixels, making them suitable for evaluating algorithms in the field of image classification and object recognition.
\end{itemize}

\subsection{ImageNet-Subsets} ImageNet~\cite{imagenet} is designed to cover a wide range of visual concepts and is commonly used for tasks such as object recognition, image classification, and object detection. It consists of millions of high-resolution labeled images spanning over thousands of object categories. Following~\cite{modelaugmentation,idc}, we evaluate our method on two subsets extracted from the ImageNet dataset. These subsets are specifically composed of 10 and 100 subclasses of ImageNet, referred to as ImageNet-10 and ImageNet-100, respectively.

\section{Baseline Description}\label{detailed_baselines}
\subsection{Coreset-Selection}
\begin{itemize}
    \item \textbf{Random}: selecting partial samples randomly from the original set.
    \item \textbf{Herding}~\cite{core_herding}: selecting data points that are close to the class centres.
    \item \textbf{K-center}~\cite{core_kcenter1,core_kcenter2}: selecting the subset using K-center algorithm, which iteratively selects centers and including points that are closest to these centers.
\end{itemize}
\subsection{Dataset-Condensation}
\myPara{Optimization-oriented.}
\begin{itemize}
    \item \textbf{DC}~\cite{dc}: matching the gradient induced by the real and synthetic images, and update the network on the condensed set.
    \item \textbf{DSA}~\cite{dsa}: applying a differentiable Siamese augmentation to images before input it to the network.
    \item \textbf{IDC}~\cite{idc}: compared to DC, IDC use a factor technique that split one image into several lower-resolution ones. Besides, IDC update the network on the original real set instead of the condensed set.
\end{itemize}
\myPara{Distribution-Matching-based.}
\begin{itemize}
    \item \textbf{CAFE}~\cite{cafe}: aligning the representation embedding of the real and synthetic images in a layer-wise manner. Moreover, CAFE utilizes a discriminant loss to enhance the discriminative properties of the condensed set.
    \item \textbf{CAFE+DSA}~\cite{cafe}: additionally applying DSA stategy to images compared to CAFE.
    \item \textbf{DM}~\cite{dm}: aligning the representation embedding of the real and synthetic images.
    \item \textbf{IDM}~\cite{idm}: applying model sampling, distribution regularization and expansion augmentation to DM~\cite{dm} to improve its performance. 
\end{itemize}

\section{More Ablation Studies}
\myPara{Compatibility with Previous DM-based Methods.}~
The proposed kernel-form loss $\mathcal{L}_{\text{M3D}}$ can be readily combined with previous DM-based methods. We substitute the loss function of DM~\cite{dm} and IDM~\cite{idm} with our $\mathcal{L}_{\text{M3D}}$, as shown in Tab.~\ref{combineDM}, the $\mathcal{L}_{\text{M3D}}$ can lead to great improvements in the performance of the examples condensed by previous DM-based methods.

\begin{table}[tb]
    \centering
    % \resizebox{0.45\textwidth}{!}{
    \begin{tabular}{l|c|c|c|c}
    \toprule
        & \multicolumn{2}{c|}{DM} &  \multicolumn{2}{c}{IDM} \\
        \midrule
       & w/o $\mathcal{L}_{\text{M3D}}$  & w/ $\mathcal{L}_{\text{M3D}}$ &  w/o $\mathcal{L}_{\text{M3D}}$  & w/ $\mathcal{L}_{\text{M3D}}$ \\
       \midrule
    IPC=10  & 48.9  & \textbf{54.7} ({$\uparrow$ 5.8}) & 58.6 & \textbf{60.5} ({$\uparrow$ 1.9})\\
    IPC=50  & 63.0  & \textbf{66.0} ({$\uparrow$ 3.0})& 67.5 & \textbf{69.6} ({$\uparrow$ 2.1})\\
         \bottomrule
    \end{tabular}
    \caption{{Compatibility of M3D with previous DM-based methods.} All experiments are conducted on CIFAR-10~\cite{cifar}.}
    % }
    \label{combineDM}
\end{table}

\section{More Visualization Results}\label{more_vis}

\begin{figure*}[tb]
  \centering
  \begin{subfigure}{0.47\linewidth}
    \includegraphics[width=\textwidth]{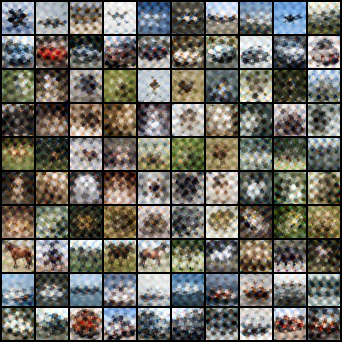}
    \caption{}
  \end{subfigure}
  \hfill
    \begin{subfigure}{0.47\linewidth}
    \includegraphics[width=\textwidth]{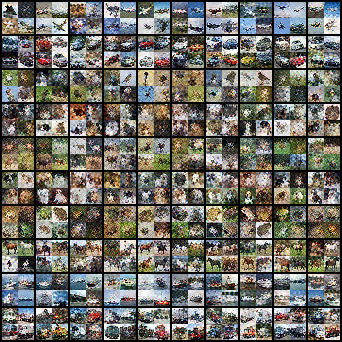}
    \caption{}
  \end{subfigure}
    \caption{{Visualization of the synthetic set of CIFAR-10 with 10 images per class.} (a) without the factor technique (b) with the factor technique.}
  \label{vis_cifar10}
\end{figure*}

\begin{figure*}[tb]
  \centering
  \begin{subfigure}{0.47\linewidth}
    \includegraphics[width=\textwidth]{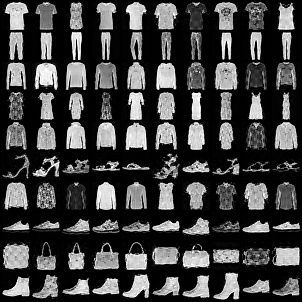}
    \caption{}
  \end{subfigure}
  \hfill
    \begin{subfigure}{0.47\linewidth}
    \includegraphics[width=\textwidth]{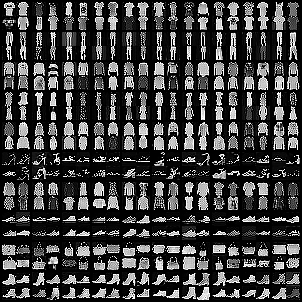}
    \caption{}
  \end{subfigure}
    \caption{{Visualization of the synthetic set of Fashion-MNIST with 10 images per class.} (a) without the factor technique (b) with the factor technique.}
  \label{vis_fmnist}
\end{figure*}

\begin{figure*}[tb]
  \centering
  \begin{subfigure}{0.47\linewidth}
    \includegraphics[width=\textwidth]{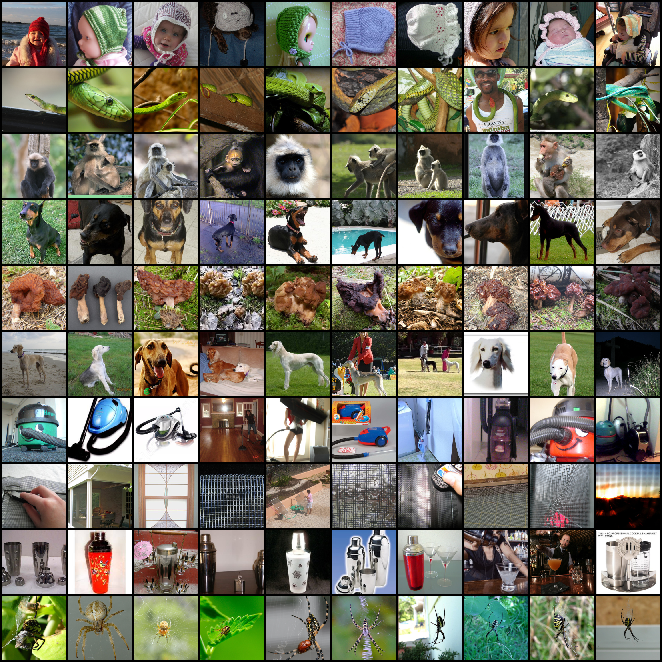}
    \caption{}
  \end{subfigure}
  \hfill
    \begin{subfigure}{0.47\linewidth}
    \includegraphics[width=\textwidth]{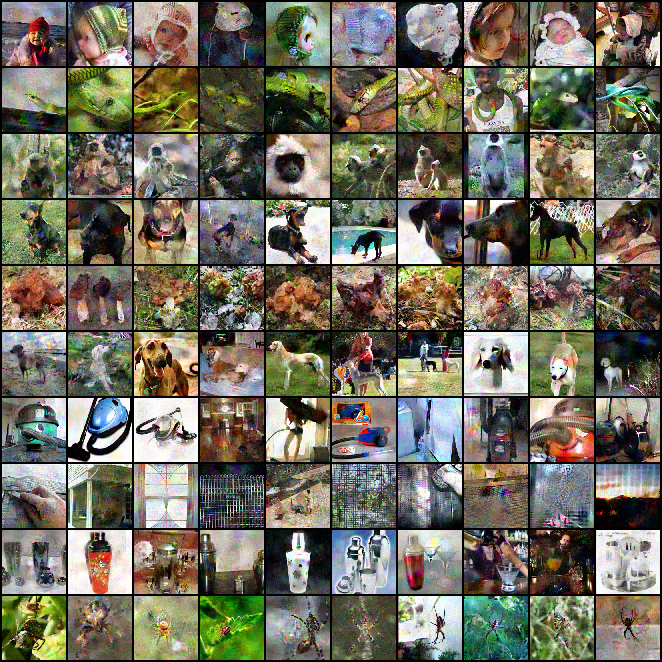}
    \caption{}
  \end{subfigure}
    \caption{{Visualization of the initialized and synthetic images of ImageNet-10 with 10 images per class.} (a) initialized images (b) synthetic images.}
  \label{vis_imgnet}
\end{figure*}

Fig.~\ref{vis_cifar10}$\sim$\ref{vis_fmnist} visualize the synthetic images condensed by our method on low-resolution datasets with and without the factor technique. We also provide the visualization of initialized and synthetic ImageNet-10 with 10 images per class in Fig.~\ref{vis_imgnet}.

\begin{figure*}[tb]
  \centering
  \begin{subfigure}{0.47\linewidth}
    \includegraphics[width=\textwidth]{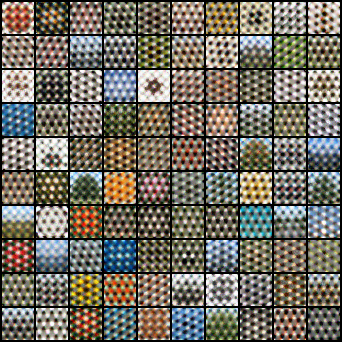}
    \caption{}
  \end{subfigure}
  \hfill
    \begin{subfigure}{0.47\linewidth}
    \includegraphics[width=\textwidth]{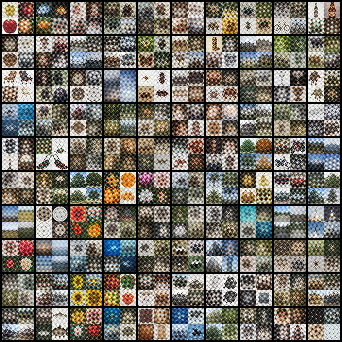}
    \caption{}
  \end{subfigure}
    \caption{{Visualization of the synthetic set of CIFAR-100 with 1 image per class.} (a) without the factor technique (b) with the factor technique.}
  \label{vis_cifar100}
\end{figure*}

\begin{figure*}[tb]
  \centering
  \begin{subfigure}{0.47\linewidth}
    \includegraphics[width=\textwidth]{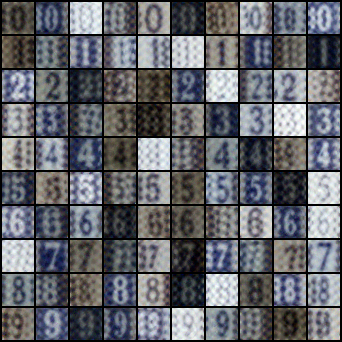}
    \caption{}
  \end{subfigure}
  \hfill
    \begin{subfigure}{0.47\linewidth}
    \includegraphics[width=\textwidth]{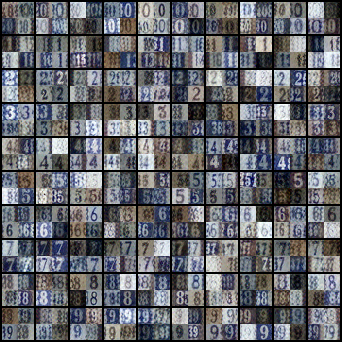}
    \caption{}
  \end{subfigure}
    \caption{{Visualization of the synthetic set of SVHN with 10 images per class.} (a) without the factor technique (b) with the factor technique.}
  \label{vis_svhn}
\end{figure*}

\end{document}